\documentclass[11pt,a4paper]{article}
\usepackage[hyperref]{acl2018}
\usepackage{times}
\usepackage{latexsym}
\usepackage{url}
\usepackage{graphicx}
\usepackage{tabularx}
\usepackage{soul}
\usepackage{xcolor}
\usepackage{booktabs}
\usepackage{paralist} %compactitem
\usepackage{enumerate}
\usepackage[normalem]{ulem}  % for underlining words
\usepackage{epstopdf}
\usepackage{xstring}

\aclfinalcopy % Uncomment this line for the final submission
\title{NRC VAD Lexicon v2: Norms for Valence, Arousal, and Dominance\\ for over 55k English Terms}

\author{Saif M. Mohammad  \\
  National Research Council Canada \\
  {\tt saif.mohammad@nrc-cnrc.gc.ca} }

\date{}

\begin{document}
\maketitle
\begin{abstract}
Factor analysis studies have shown that the primary dimensions of word meaning 
are \textit{Valence (V)}, \textit{Arousal (A)}, and \textit{Dominance (D)} (also referred to in social cognition research as \textit{Competence (C)}). 
% Social psychologists have shown that \textit{Competence (C)} (along with \textit{Warmth (W)}) are the primary dimensions along which we assess other people and groups. 
These dimensions impact various aspects of our lives from social competence and emotion regulation to success in the work place and how we view the world.
We present here the NRC VAD Lexicon v2, which has human ratings of valence, arousal, and dominance for more than 55,000 English words and phrases. Notably, it adds entries for $\sim$25k additional words to v1.0. It also now includes for the first time entries for common multi-word phrases ($\sim$10k). 
We show that the associations are highly reliable.
% split-half reliability of 0.82 Spearman and 0.89 Pearson correlation. 
% We use the lexicon to study the rate at which children acquire VAD words with age. 
% Finally, we show that 
The lexicon enables a wide variety of  %bias and stereotype 
research in psychology, NLP, public health, digital humanities, and social sciences. 
The NRC VAD Lexicon v2 is made freely available for research 
through our project webpage.\footnote{http://saifmohammad.com/WebPages/nrc-vad.html}
% through case studies on various target entities.
\end{abstract}

% Words play a central role in language and thought. Factor analysis studies have shown that the primary dimensions of meaning %, especially for affective words, 
% are valence, arousal, and dominance (VAD). 
% % In this paper, 
% % We describe how 
% We obtain reliable human ratings of valence, arousal, and dominance for more than 20,000 English words. We use best--worst scaling to obtain fine-grained scores and address issues of annotation consistency that plague traditional rating scale methods of annotation.  
% We show that the ratings obtained are vastly more reliable than those in existing lexicons.
% % We conduct an extensive demographic analysis and 
% We also show that there exist %s small but 
% statistically significant differences in 
% % agreements 
% the shared understanding of valence, arousal, and dominance across demographic variables such as age, gender, and personality. % traits. 

% Words of Warmth (and its translations to over 100 languages) is freely available.\\[3pt] 
% \url{http://saifmohammad.com/Words of Warmth.html}

\section{Introduction}

Words are an immensely expressive medium to convey meaning and intent. They play a central role in our  understanding and descriptions of the world around us. 
Thus words are powerful mediums for understanding language, people, thought, behaviour, and emotions.
% Some believe that the structure of a language even affects how we think (principle of linguistic relativity aka the Sapir–Whorf hypothesis). 
Several influential factor analysis studies have shown 
that the three most important, largely independent, dimensions of word meaning are
valence (positiveness--negativeness/pleasure--displeasure), arousal (active--passive), and dominance aka competenece (dominant--submissive, competent--incompetent, powerful--weak, in control--out of control)
\cite{Osgood1957,russell1980circumplex,russell2003core}. (We will refer to the three dimensions individually as {\it V, A,} and \textit{D}, and together as \textit{VAD}.)
Thus, when comparing the meanings of two words, we can compare their degrees of V, A, and D. 
For example, the word {\it banquet} indicates more positiveness  than the word {\it funeral}; {\it nervous} indicates  more arousal than {\it lazy}; and {\it fight} indicates more dominance than  {\it delicate}.

Access to these degrees of valence, arousal, and dominance of words is beneficial for a number of applications, including those in natural language processing (e.g., automatic sentiment and emotion analysis of text), in cognitive science (e.g., for understanding how humans represent and use language), in psychology (e.g., for understanding how people view the world around them), in social sciences (e.g., for understanding relationships between people), and even in evolutionary linguistics (e.g., for understanding how language and behaviour inter-relate to give us an advantage). 

% Existing VAD lexicons \cite{bradley1999affective,warriner2013norms}

\noindent {\bf Existing Affect Lexicons:} The \newcite{bradley1999affective} 
% created the first lexicon of 
% annotated 
lexicon has more than 1000 words 
% that were manually annotated for VAD. 
with real-valued scores of 
valence, arousal, and dominance. 
 For each word, they
 asked annotators to rate valence, arousal, and dominance---for more than 1,000 words---on a 9-point rating scale. The ratings from multiple annotators were averaged to obtain a score between 1 (lowest V, A, or D) to 9 (highest V, A, or D).
Their lexicon, called the {\it Affective Norms of English Words (ANEW)}, has since been widely used across many different fields of study.
ANEW was also translated into non-English languages: e.g., \newcite{moors2013norms} for Dutch,  \newcite{vo2009berlin} for German, and \newcite{redondo2007spanish} for Spanish.
\newcite{warriner2013norms} created a VAD lexicon for more than 13,000 words, using a similar annotation method as for ANEW. % as was used for ANEW.
The NRC VAD lexicon v1.0 \cite{mohammad-2018-obtaining} is the largest manually created VAD lexicon (in any language), and the only one that was created via comparative annotations (instead of rating scales). It has entries for about 20,000 English words.

\noindent {\bf Our Work:} In this paper, we describe how we substantially add to the NRC VAD lexicon v1.0 by obtained human ratings of valence, arousal, and dominance for about 25,000 additional English words and more than 10,000 common multi-word expressions (MWEs). 
% The words include commonly used English terms, terms commonly found in English tweets, and terms from existing affect word lists such as the General Inquirer \cite{Stone66}, ANEW \cite{bradley1999affective}, and the NRC Emotion Lexicon \cite{MohammadT10,MohammadT13}.
The scores are fine-grained real-valued numbers from -1 (lowest V, A, or D) to 1 (highest V, A, or D). 
We show that the annotations lead to reliable VAD score (split-half reliability scores of  $r=0.99$ for valence, $r=0.98$ for arousal, and $r=0.96$ for dominance.) 
We will refer to this lexicon with about 55k entries as the {\it NRC Valence, Arousal, and Dominance (VAD) Lexicon v2}. 

% Measuring correlations ($r$) on repeat annotations, through metrics such as {\it split-half reliability (SHR)}, is a common method to evaluate the reliability of annotations.

All of the annotation tasks described in this paper were approved by 
% the National Research Council Canada's Institutional 
our institution's review board, which examined the methods to ensure that they were ethical. Special attention was paid to obtaining informed consent and protecting participant anonymity. 
The NRC VAD Lexicon v2 is made freely available for research 
through our project webpage.\footnote{http://saifmohammad.com/WebPages/nrc-vad.html}

\section{Related Work}

\noindent {\bf Primary Dimensions of Meaning and Affect:} \newcite{Osgood1957} asked human participants to rate words along dimensions
of opposites such as {\it heavy--light, good--bad, strong--weak,} etc.
Factor analysis of these judgments revealed 
that the three most prominent dimensions of meaning
are evaluation ({\it good--bad}), potency ({\it strong--weak}), and activity ({\it active--passive}). 
\newcite{russell1980circumplex,russell2003core} showed through similar analyses of emotion words that the three primary independent dimensions of emotions are valence or pleasure (positiveness--negativeness/pleasure--displeasure), arousal (active--passive), and dominance (dominant--submissive). % According to the VAD model of affect, 
They argue that individual emotions such as joy, anger, and fear are points in a three-dimensional space of valence, arousal, and dominance.
It is worth noting that even though the names given 
% to the three dimensions 
by Osgood et al.\@ and Russel et al.\@ are different, they describe similar dimensions \cite{bakker2014pleasure}.

\noindent {\bf Primary Dimensions of Social Cognition:}
In a similar vein, Social Psychology research has shown that \textit{warmth (W)}  (friendliness, trustworthiness, and sociability) and \textit{competence (C)} (ability, power, dominance, and assertiveness) are core dimensions of social cognition and stereotypes \cite{fiske2002,bodenhausen2012social,fiske2018stereotype,abele2016facets,koch2024validating}.
  That is, human beings quickly and subconsciously judge (assess) other people, groups of people, and even their own selves along the dimensions of warmth and competence (likely because of evolutionary pressures \cite{macdonald1992warmth,eisenbruch2022warmth}. Assessing W and C was central to early human survival (e.g., to anticipate whether someone will help them build useful things or whether they might steal their resources).
  Note that even though the Social Psychology work prefers the term competence, it essentially refers to the same dimension that in Affective Science research is called dominance. 
  % Warmth on the other hand is considered to be a subset of valence. 
  Warmth is considered to be a primary component of valence, which in turn is evolutionarily central to the approach--avoid response, and so some researchers argue that the ability to assess of warmth emerges earlier than dominance/competence \cite{cuddy2007bias}. This is the \textit{primacy of valence} hypothesis.

The dimensions of W and C (or D) have been shown to have substantial implications on a wide variety of facets, including: interpersonal status \cite{swencionis2017warmth}, social class \cite{durante2017social}, self-beliefs \cite{wojciszke2009two}, political perception \cite{fiske2014never}, child development \cite{ROUSSOS2016133},  cultural analyses \cite{fiske2016stereotype}, as well as
professional and organizational outcomes, such as hiring, employee evaluation, and allocation of tasks and resources \cite{cuddy2011dynamics}.

% Psychologists have argued that  some emotions are more basic than others  \cite{Ekman92,Plutchik80,Parrot01,frijda1988laws}.\footnote{However, 
% they disagree on which emotions (and how many) should be classified as basic emotions---some propose 6, some 8, some 20, and so on.}
% Thus, most work on capturing word--emotion associations has focused 
% on a handful of emotions, especially since
% manually annotating for a large number of emotions is arduous. 
% In this project, the goal is to create an affect intensity lexicon for the eight emotions emotions: anger, fear, joy, sadness, disgust, trust, anticipation, and surprise.
% These are the eight emotions considered to be most basic by \cite{Plutchik80}. The eight emotions include the six emotions considered most basic by \cite{Ekman92},
% as well as trust and anticipation.

\noindent {\bf Existing Affect Lexicons:} We already discussed several VAD lexicons in the Introduction such as \newcite{bradley1999affective}, \newcite{warriner2013norms}, \newcite{moors2013norms}, \newcite{vo2009berlin}, \newcite{redondo2007spanish}, and \newcite{mohammad-2018-obtaining}. Other work has focused on creating sentiment lexicons, where words are marked for whether they denotate or connotate 
% positive or negative 
sentiment (but do not include fine valence scores or any information about arousal and dominance).  
Examples of such lexicons include the General Inquirer \cite{Stone66}, MPQA \cite{Wiebe05}, and the NRC Emotion Lexicon \cite{MohammadT13,MohammadT10}.
The NRC Emotion Lexicon also includes entries for whether a words is associated with any of the eight basic motions \newcite{Plutchik80}.
% The NRC Emotion Lexicon was created by crowdsourcing and it includes entries for
% about 14,000 words and eight Plutchik emotions 
%\cite{MohammadT13,MohammadT10}.\footnote{http://www.purl.org/net/saif.mohammad/research}

The NRC Emotion Intensity Lexicon \cite{LREC18-AIL} has real-valued scores of intensity for the words in the NRC Emotion Lexicon associated with any of the eight emotions: anger, anticipation, disgust, fear, joy, sadness, surprise, and trust. 
The NRC WorryWords Lexicon \cite{worrywords-emnlp2024} is a list of over 44,000 English words and real-valued scores indicating their associations with anxiety: from -3 (maximum calmness) to 3 (maximum anxiety).

\noindent {\bf Automatically Creating Affect Lexicons:} There is growing work on automatically determining word--sentiment and word--emotion associations. These include
\newcite{COIN:COIN12024,Mohammad12,StrapparavaV04,yang2007building,yu2015predicting,staiano2014depechemood,bandhakavi2021emotion,muhammad-etal-2023-afrisenti} to name just a few.  These methods  often assign a real-valued score  representing the degree of association. 
The VAD Lexicon can be used to evaluate how accurately the automatic methods capture valence, arousal, and dominance.

% All of the emotion work and a vast majority of the valence (sentiment) work has used categorical annotation or a coarse rating scale to obtain annotations.
% This is not surprising, because it is difficult for humans to provide direct scores at a fine granularity.   A common problem is inconsistencies in annotations among different annotators. 
% One annotator might assign a score of 7.9 to a word, whereas another annotator may assign a score of 6.2 to the same word. 
% It is also common that the same annotator assigns different scores to the same word 
% at different points in time.
% Further, annotators often have a bias towards different parts of the scale, known as {\it scale region bias}.
% Despite this,  a key question is whether humans are able to distinguish affect at only four or five coarse levels, or whether we can discriminate across much smaller affect intensity differences.

\section{Obtaining Human Ratings of Valence, Arousal, and Dominance}
The keys steps in obtaining the new annotations were as follows: 
\begin{compactenum}
\item selecting the terms to be annotated 
\item developing the questionnaire
\item developing measures for quality control (QC)
\item annotating terms on a crowdsource platform
\item discarding data from outlier annotators (QC)
\item aggregating data from multiple annotators to determine the VAD association scores
% for each of the terms. 
\end{compactenum}
We describe each of the steps below.\\[3pt]
\noindent {\bf 1. Term Selection.}
The NRC VAD Lexicon v1.0 already included a large number of common English words from many different sources: 
% Specifically, we include terms from the following sources:\\[-18pt]
\begin{itemize}
\item All terms in the NRC Emotion Lexicon \cite{MohammadT13,MohammadT10}. 
The NRC Emotion Lexicon has about 14,000 words that are annotated to indicate whether they are associated with any of the eight basic emotions: 
% identified by \newcite{Plutchik62} 
anger, anticipation, disgust, fear, joy, sadness, surprise, and trust \cite{Plutchik80}
The NRC lexicon terms were in turn chosen by taking the content words that occur frequently in the Google n-gram corpus \cite{BrantsF06}.\\[-19pt]
\item All 4,206 terms in the positive and negative lists of the General Inquirer \cite{Stone66}.\\[-19pt]
\item All 1,061 terms listed in ANEW \cite{bradley1999affective}.\\[-19pt]
\item All 13,915 terms listed in the \newcite{warriner2013norms} lexicon.\\[-19pt]
\item 520 words from the Roget's Thesaurus categories corresponding to the eight basic Plutchik emotions.\footnote{http://www.gutenberg.org/ebooks/10681}\\[-19pt]
\item About 1000 high-frequency content terms, including emoticons, from the Hashtag Emotion Corpus (HEC)  \cite{Mohammad12}. All tweets in the The HEC include at least one of the eight basic emotion words as a hashtag word (e.g., {\it \#anger, \#sadness}, etc.). 
% Note that this set of terms includes both terms that are more common in social media communication (for example, {\it :), soannoyed, grrrrr, stfu}, and {\it thx})
% as well as regular English words.\footnote{Some of the terms included from the Twitter source were deliberate spelling variations of English words, for example, {\it bluddy} and {\it sux}.}\\[-22pt]
\end{itemize}
\noindent The union of all of the above sets resulted in about 20k terms that were then annotated for valence, arousal, and dominance.

To add new terms we wanted again focus on common English terms not included in v1.0, and in addition, we wanted to include common phrases (multi-word expressions, light verb constructions, etc.).
Finally, we wanted to include terms for which other linguistically interesting annotations already exists (such as concreteness and age of acquisition ratings).
Therefore we included terms from the Prevalence dataset \cite{brysbaert2019word}. 
%    that are marked as known to 70\% or more respondents.
    This dataset has prevalence scores (how widely a word is known by English speakers), determined directly by asking people, for 62,000 lemmas. We included a term if it was marked as known to at least 70\% of the people who provided responses for the term. 
    % There were 45,222 such terms in total.
    (From this set we removed terms that are common person names or city names.)
This resulted in close to 25k unigrams.    
We also included $\sim$10.5k most common multi-word expressions from the \newcite{muraki2023concreteness} dataset. This dataset has concreteness ratings for about 62k English MWEs, as well as their frequencies in a subtitles corpus \cite{brysbaert2012adding}.

% (specifically, words marked as being known to over 70\% of the respondents). 
% Summary details of each of these is provided in the Appendix.

\noindent{\bf 2. VAD Questionnaires}
The questionnaires used to annotate the data 
 were developed  after several rounds of 
 pilot annotations. 
 Detailed directions, including notes directing respondents to consider predominant word sense (in case the word is ambiguous) and example questions (with suitable responses) were provided. (See Appendix.)
 The primary instruction and the questions presented to annotators are shown below.\\[-17pt]
% The annotation questions and the instructions for the annotators are shown in a supplementary file. 

{
\noindent\makebox[\linewidth]{\rule{0.48\textwidth}{0.4pt}}\\% [-8pt]
{ \small
% Summary Instructions
\noindent VALENCE: Consider positive feelings (or positive sentiment)  to be a broad category that includes:\\
\indent \textit{positiveness / pleasure / goodness / happiness /\\
\indent greatness / brilliance / superiority / health  etc.}\\
Consider negative feelings (or negative sentiment) to be a category that includes:\\
\indent \textit{negativeness / displeasure /badness / unhappiness /\\ \indent insignificance / terribleness / inferiority / sickness etc.}\\
% nonchalant, uninterested, 
 If you do not know the meaning of a word or are unsure, you can look it up in a dictionary (e.g., the Merriam Webster) or on the internet.

\noindent Quality Control

 Some questions have pre-determined correct answers. If you mark these questions incorrectly, we will give you immediate feedback in a pop-up box. An occasional misanswer is okay. However, if the rate of misanswering is high (e.g., $>$20\%), then all of one's HITs may be rejected.

\noindent Select the options that most English speakers will agree with.\\[4pt]
\noindent Q1.  $<$term$>$ is often associated with:\\[-1pt]
\indent 3: very positive feelings\\[-1pt]
\indent 2: moderately positive feelings\\[-1pt]
\indent 1: slightly positive feelings\\[-1pt]
\indent 0: not associated with positive or negative feelings\\[-1pt]
\hspace*{-2mm} \indent -1: slightly negative feelings\\
\hspace*{-2mm} \indent -2: moderately negative feelings\\
\hspace*{-2mm} \indent -3: very negative feelings\\[-8pt]
}
\noindent\makebox[\linewidth]{\rule{0.48\textwidth}{0.4pt}}\\[-20pt]

}

{
\noindent\makebox[\linewidth]{\rule{0.48\textwidth}{0.4pt}}\\% [-8pt]
{ \small
\noindent AROUSAL: This task is about words and their association with activeness or arousal.
\noindent Consider activeness or arousal to be a broad category that includes:

    \textit{active, aroused, stimulated, %frenzied, 
    excited, jittery, alert,} etc.

\noindent Consider inactiveness or calmness to be a broad category that includes:

    \textit{inactive, calm, unaroused, passive, relaxed, sluggish,} etc.

\noindent This task is not about sentiment. (For example, something can be positive and inactive (such as flower), positive and active (such as exercise and party), negative and active (such as murderer), and negative and inactive (such as negligent).\\[-16pt]

\noindent\makebox[\linewidth]{\rule{0.48\textwidth}{0.4pt}}\\[-16pt]

}

{
\noindent\makebox[\linewidth]{\rule{0.48\textwidth}{0.4pt}}\\% [-8pt]
{ \small
% Summary Instructions
\noindent DOMINANCE: This task is about words and their association with dominance, competence, control of situation, or powerfulness.
Consider dominance, competence, control of situation, or powerfulness to be a broad category that includes:

   \indent \textit{ dominant, competent, in control of the situation,}\\ \indent \textit{powerful, influential, important, autonomous,} etc.\\
\noindent Consider submissiveness, incompetence, controlled by outside factors, or weakness to be a broad category that includes:

    \indent \textit{submissive, incompetent, not in control of the situation,}\\
    \indent \textit{weak, influenced, cared-for, guided,} etc.\\
\noindent This task is not about sentiment. (For example, something can be positive and weak (such as a flower petal) and something can be negative and strong (such as tyrant).\\[-16pt]

% \noindent If you do not know the meaning of a word or are unsure, you can look it up in a dictionary (e.g., the Merriam Webster) or on the internet.

% \noindent A rule of thumb is that a term associated with more dominance, competence, control of situation, or powerfulness tends to often occur in sentences that convey dominance, competence, control of situation, or powerfulness, whereas a term associated with more submissiveness, incompetence, controlled by outside factors, or weakness tends to often occur in sentences that convey submissiveness, incompetence, controlled by outside factors, or weakness.

\noindent\makebox[\linewidth]{\rule{0.48\textwidth}{0.4pt}} 

}

% \noindent The full questionnaire will be made available on the project webpage. 

\noindent{\bf 3. Quality Control Measures.}
%brb author no s below
About 2\% of the data was annotated beforehand by the authors and interspersed with the rest. These questions are referred to as \textit{gold} (aka \textit{control}) questions. 
% During crowd annotation, 
% We interspersed the gold questions with the other questions.
% and the annotator is not aware which is which. However, 
Half of the gold questions were used to provide immediate feedback to the annotator (in the form of a pop-up on the screen) in case they mark them incorrectly. We refer to these as \textit{popup gold}. This helps prevent the situation where one annotates a large number of instances without realizing that they are doing so incorrectly. 
It is possible, 
that some annotators share answers to gold questions with each other (despite this being against the terms of annotation). 
% The gold questions also served as examples to guide the annotators.
Thus, the other half of the gold questions were also separately used to track how well an annotator was doing the task, but for these gold questions no popup was displayed in case of errors. 
We refer to these as 
\textit{no-popup gold}.\\[-13pt]
% If a crowd worker answered a gold question incorrectly, then they were immediately notified.
% \sm{the annotation was discarded, and an additional annotation was requested from a different annotator}. 

\noindent{\bf 4. Crowdsourcing.} 
We setup the annotation tasks on the crowdsourcing platform, {\it Mechanical Turk}.
 In the task settings, we specified that we needed annotations from nine people for each word. 
We obtained annotations from native speakers of English residing around the world. Annotators were free to provide responses to as many terms as they wished. 
The annotation task was approved by 
% the National Research Council Canada's Institutional 
our institution's review board.
% , which reviewed the proposed methods to ensure that they were ethical.
% We used CrowdFlower's gold annotations scheme for quality control, wherein 

\noindent {\it Demographics:} About 95\% of the respondents who annotated the words live in USA. The rest were from India, United Kingdom, and Canada. 
The average age of the respondents was 34 years. Among those that disclosed their gender, about 53\% were female, 47\% were male.\footnote{Respondents were shown optional text boxes to disclose their demographic information as they choose; especially important for social constructs such as gender, in order to give agency to the respondents and to avoid binary language.}\\[-13pt]

\noindent{\bf 5. Filtering.} % and Re-annotation} %  as part of Quality Control} 
If an annotator's accuracy on the gold questions (popup or non-popup) fell below 80\%, then they were refused further annotation, 
 and all of their annotations were discarded (despite being paid for).
% We then obtained fresh annotations for those terms from other annotators.
% This served as a mechanism to avoid malicious and random annotations.
  % However, because of the way the gold questions work in CrowdFlower, they were annotated by more than six people. 
%  Both the minimum and the median number of annotations per item was 10. 
 See Table~\ref{tab:ann} for summary statistics.\\[-13pt]

% brb
% warmth-unigrams-lexicon.twb
%  \begin{figure*}[t]
% 	     \centering
% 	     \includegraphics[width=0.9\textwidth]{figures/warmth-distribution.png}
% 	    \vspace*{-3mm}
%       \caption{Distribution of terms in Words of Warmth: percentage and number of terms associated with each class.}
% 	     \label{fig:Words of Warmth-distrib}
% \vspace*{-3mm}
% 	 \end{figure*}

\noindent{\bf 6. Aggregation.} 
Every response was mapped to an integer from -3 (highly negative/inactive/submissive) to 3 (highly positive/active/dominant) as follows: 
\begin{compactitem}
    \item highly positive/active/dominant: 3
    \item moderately positive/active/dominant: 2
    \item slightly positive/active/dominant: 1
    \item neither positive/active/dominant nor  negative/inactive/submissive: 0
    \item slightly negative/inactive/submissive: -1
    \item moderately negative/inactive/submissive: -2
    \item highly negative/inactive/submissive: -3
\end{compactitem}
The final  score for each term is simply the average score it received from each of the annotators.
 The scores were then linearly transformed to the interval: -1 (highest negativeness/inactivity/submissiveness) 
 to 1 (highest positiveness/activity/dominance).
% The scores are linearly transformed into the interval -1 (lowest V/A/D) to 1 (the highest V/A/D).
 
% Since degree of emotion is a unipolar scale, 
% We linearly transform the -1 to 1 scores to scores in the range 0 (least emotion intensity) to 1 (the most emotion intensity).
% We also created a categorical version of the sociability (S) lexicon by labeling all words that got an average score $\geq 2.5$ as \textit{high}, $\geq 1.5$ and $< 2.5$ as \textit{moderate}, $\geq 0.5$ and $< 1.5$ as \textit{slight },
% $> -0.5$ and $< 0.5$ as \textit{neither sociable nor unsociability}, and so on. The categorical version of the trust (T) lexicon was created similarly.

The terms and their VAD scores were added to the NRC VAD Lexicon v1 to create the NRC VAD Lexicon v2.
% We refer to the
% list of words
% along with their 
% % real-valued final 
% scores and categorical labels for WST as the {\it Words of Warmth Lexicons}. 

%brb
% Table \ref{tab:examples} in the Appendix shows example entries from the lexicons. % with the highest and lowest scores.
% Figure \ref{fig:Words of Warmth-distrib} shows the distribution of the different classes. 

% questionnaires that we used, and the crowdsourced annotations.
% to
% obtain the VAD ratings.
% created the % Valence, Arousal, and Dominance 
% VAD Lexicon.

\begin{table}[t!]
\begin{center}
%\vspace*{-4mm}
\small{
\begin{tabular}{lrrr}
\hline 
{\bf Version} 	             & \bf \#Words& \bf \#MWEs & \bf \#Total\\\hline
v1.1 (2018)     & 19,839       & 132     & 19,971\\
% v2.1 (2025)     & 44,728       &10,073 & 54,801\\
v2.1 (2025)     & 44,928       &10,073 & 55,001\\
\hline
\end{tabular}
}
\vspace*{-2mm}
\caption{\label{tab:ann} {Number of terms in the NRC VAD Lexicon in version 1.1 and 2.1.}
}
% \vspace*{-3mm}
\end{center}
\end{table}

\begin{table}[t!]
\begin{center}
%\vspace*{-4mm}
\small{
\begin{tabular}{lrrr}
\hline 
{\bf Version} 	             & \bf Avg.\@ \#Annot. &\bf SHR ($\rho$)		 &\bf SHR ($r$) \\\hline
valence         & 7.83      &0.98   &0.99     \\
arousal         & 7.96      &0.97       &0.98     \\
dominance       & 8.06      &0.96     &0.96\\
\hline
\end{tabular}
}
\vspace*{-2mm}
\caption{\label{tab:shr-words} {Average number of annotations per word and split half reliability measured through both Spearman rank ($\rho$) and Pearson's ($r$) correlations. Scores in the 0.9s indicate high reliability.}
}
% \vspace*{-3mm}
\end{center}
\end{table}

\section{Reliability of the Annotations} 

A useful measure of quality is the reproducibility of the end result---repeated independent manual
annotations from multiple respondents should result in similar  scores.
To assess this reproducibility, we calculate
average {\it split-half reliability (SHR)} over 1000 trials. SHR is a common way to determine reliability of responses to generate scores on an ordinal scale %in the fields of psychology and psycholinguistics 
\cite{weir2005quantifying}.} 
% SHR a commonly used approach to
% determine consistency in psychological studies, that we employ as follows.   
All annotations for an
item are randomly split into two halves. Two separate sets of scores are aggregated, just as described in Section 3 (bullet 6), from the two halves. 
% but independently from the two SHR halves.  
% Then the correlation between the two sets of scores is calculated. 
Then we determine how close the two sets of scores are (using a metric of correlation). % using Spearman Rank Correlation. 
This is repeated 1000 times and the correlations are averaged.
The last two columns in Table~\ref{tab:shr-words} show the results (split half-reliabilities). Spearman rank and Pearson correlation scores of over 0.95  for V, A, and D indicate high reliability of the real-valued scores obtained from the annotations. (For reference, if the annotations were random, then repeat annotations would have led to an SHR of 0. Perfectly consistent repeated annotations lead to an SHR of 1. Also, similar past work on word--anxiety associations had SHR scores in the 0.8s \cite{mohammad-2024-worrywords}.)

\section{Applications and Future Work}

The large number of entries in the VAD Lexicon and the high reliability of the scores make it useful for a number of research projects and applications. We list a few below:
\begin{itemize}
\item  Understanding valence, arousal, and dominance, and the underlying mechanisms; how VAD relate to our mind and body; how VAD change with age, socio-economic status, weather, green spaces, etc.\\[-20pt]
\item Determining how VAD manifest in language; how language shapes our VAD; how culture shapes the language of VAD; etc.\\[-20pt]
\item Tracking the degree of VAD towards targets of interest such as climate change, %BB commercial products, 
government policies, biological vectors, etc.\\[-20pt] %; tracking common targets of anxiety;  
\item Studying stereotypes and social cognition; using the dominance aka competence lexicon to study how competence assessment capabilities develop in children and to track perceptions of competence towards various targets of interest.\\[-20pt]
\item Developing automatic systems for detecting VAD; To provide features for automatic sentiment or emotion detection systems. They can also be used to obtain sentiment-specific word embeddings and sentiment-specific sentence representations.\\[-20pt] 
%  WASSA-2017 shared task on inferring the intensity of emotion felt by a person based on their tweet \cite{MohammadB17wassa}. 
\item To study the  interplay between the categorical emotion model and the VAD model of affect. Much of the prior work has only explored one of the two models. The VAD lexicon can be used along with lists of words associated with emotions such as joy, sadness, fear, etc. to study the correlation of V, A, and D, with those emotions.\\[-20pt] 
% \item We will use the lexicon to identify syllables and phonemes that are associated with high V/A/D, that is, syllables that tend to occur more often than average in high V/A/D words. 
\item To identify syllables that consistently tend to occur in words with high VAD scores. This has implications in understanding how some syllables and sounds have a tendency to occur in words referring to semantically related concepts. Identifying V, A, and D scores associated with syllables is also useful in generating names for literary characters and commercial products that have the desired affectual response.    \\[-20pt] 
\item Studying VAD in story telling; its relationship with central elements of narratology such as conflict and resilience.
To identify high V, A, and D words in books and literature. To facilitate work of researchers in digital humanities. To facilitate work on literary analysis.\\[-20pt]
\item As a source of gold (reference) scores, the entries in the VAD lexicon can be used in the evaluation of automatic methods of determining V, A, and D.\\[-20pt] 
\item The dataset is also of potential use to psychologists and evolutionary linguists interested in determining how evolution shaped the representation of the world around us, and why certain personality traits are associated with higher or lower shared understanding of valence, arousal, and dominance of words.\\[-16pt]
\end{itemize}
\noindent Apart from exploring the applications above, we are also interested in creating VAD lexicons for other languages, especially Chinese, Hindi, Arabic, Spanish, and German. We can then explore characteristics of valence, arousal, and dominance that are common across cultures.

\section{Conclusions}
We present here the NRC VAD Lexicon v2, which has human ratings of valence, arousal, and dominance for more 
% than 44,000 English words and 10,000 multi-word expressions. 
55,000 English terms. Compared to v1, it has entries for an additional $\sim$25k words. It also now includes for the first time entries for common multi-word expressions ($\sim$10k).
We provide a detailed description of how the terms were selected, the annotation process, and various measures for quality control.
We show that the ratings are highly reliable
(split-half reliability of over 0.95 for all three dimensions). 
% We use the lexicon to study the rate at which children acquire VAD words with age. 
% Finally, we show that 
The lexicon enables a wide variety of  %bias and stereotype 
research in psychology, NLP, public health, digital humanities, and social sciences. 
% The NRC VAD Lexicon v2.0 
It is made freely available for research 
through our project webpage.\footnote{http://saifmohammad.com/WebPages/nrc-vad.html}

\section{Limitations}
\label{sec:limitations}

The lexicon created is one of the largest that exist with wide coverage and a large number of annotators (thousands of people as opposed to just a handful). 
% to quantify the emotions expressed in utterances are also limited in their 
However, no lexicon can cover the full range of linguistic and cultural diversity in emotion expression. 
% they do not cover a comprehensive set of morphological variants of words, 
% do not disambiguate between word senses (considering only the dominant sense), 
% capture word--emotion associations as perceived (annotated) by a small sub-section of the world population, 
The lexicons are largely restricted to words that are most commonly used in Standard American English and they capture emotion associations as judged by American native speakers of English. 
% This work develops the VAD lexicon for English, based on responses primarily from people residing in the US. Thus it is important to contextualize any conclusions as those applying to US English speakers. 
Annotators on Mechanical Turk are not representative of the wider US population. However, obtaining annotations from a large number of annotators (as we do) makes the lexicon more resilient to individual biases and captures more diversity in beliefs.
We see this work as a first step that paves the way for more work using responses from various other groups of people and in various other languages. 
See \citet{Mohammad23ethicslex} for a detailed discussion of the limitations and best-practices in the use of emotion lexicons.

% \noindent See discussions of limitations in how the lexicons can be used and interpreted in the Ethics Statement below (\S \ref{sec:ethics}).

\section{Ethics and Data Statement}
\label{sec:ethics}

The crowd-sourced task presented in this paper was approved by our Institutional Research Ethics Board. 
Our annotation process stored no information about annotator identity and as such there is no privacy risk to them.  The individual words selected did not pose any risks beyond the risks of occasionally reading text on the internet. 
The annotators were free to do as many word annotations as they wished. The instructions included a brief description of the purpose of the task (Figures \ref{fig:val-q} through \ref{fig:dom-ex}}).

VAD assessments are complex, nuanced, and often instantaneous mental judgments. Additionally, each individual may use language to convey these assessments slightly differently.
% See \citet{Mohammad23ethicslex} for a discussion of good practices and ethical considerations when using emotion lexicons. See \citet{Mohammad22AER} for a broader discussion of ethical considerations relevant to automatic emotion recognition.
See \citet{Mohammad23ethicslex} for a detailed discussion of ethical considerations when computationally analyzing emotions and VAD using emotion lexicons. 
We discuss below some of the notable considerations. 
 (See \citet{Mohammad22AER} for a broader discussion of ethical considerations relevant to automatic emotion recognition.)
\begin{compactenum}
    \item \textit{Coverage:} We sampled a large number of English words from other lexical sources (which themselves sample from many sources). Yet, the words included do not cover all domains, genres, and people of different locations, socio-economic strata, etc.\@ equally. It likely includes more of the vocabulary common in the United States with socio-economic and educational backgrounds that allow for technology access.
 
    \item \textit{Word Senses and Dominant Sense Priors:} Words when used in different senses and contexts may be associated with different degrees of VAD associations. The entries in in the VAD Lexicon are indicative of the associations with the predominant senses of the words. This is usually not problematic because most words have a highly dominant main sense (which occurs much more frequently than the other senses). 
In specialized domains, some terms might have a different dominant sense than in general usage. Entries in the lexicon for such terms should be appropriately updated or removed. 
Further, any conclusions using the lexicon should be made based on relative change of associations using a large number of textual tokens. For example, if there is a marked increase in low-valence words from one period to the next, where each period has thousands of word tokens, then the impact of word sense ambiguity is minimal, and it is likely that some broader phenomenon is causing the marked increase in low-valence words. (See last two bullets.)

    \item \textit{Not Immutable:} The VAD scores do not indicate an inherent unchangeable attribute. The associations can change with time (e.g., the decrease in negativeness associated with \textit{inter-race relationships} over the last 100 years), but the lexicon entries are fixed. They pertain to the time they are created. However, they can be updated with time.
  
    \item \textit{Socio-Cultural Biases:} The annotations for VAD capture various human biases. These biases may be systematically different for different socio-cultural groups. Our data was annotated by mostly US, Canadian, UK, and Indian English speakers, but even within these countries there are many diverse socio-cultural groups.
    Notably, crowd annotators on Amazon Mechanical Turk do not reflect populations at large. In the US for example, they tend to skew towards male, white, and younger people. However, compared to studies that involve just a handful of annotators, crowd annotations benefit from drawing on hundreds and thousands of annotators (such as this work). 
 
    \item \textit{Inappropriate Biases:} Our biases impact how we view the world, and some of the biases of an individual may be inappropriate. For example, one may have race or gender-related biases that may percolate subtly into one's notions of VAD associated with words. 
    Our dataset curation was careful to avoid words from problematic sources. We also ask people annotate terms based on what most English speakers think (as opposed to what they themselves think). This helps to some extent, but the lexicon may still capture some historical VAD associations with certain identity groups. This can  be useful for some socio-cultural studies; but we also caution that VAD associations with identity groups be carefully contextualized to avoid false conclusions.    
  
    \item \textit{Perceptions (not “right” or “correct” labels):} Our goal here was to identify common perceptions of WTS association. These are not meant to be ``correct'' or ``right'' answers, but rather what the majority of the annotators believe based on their intuitions of the English language.
   
    \item \textit{Avoid Essentialism:} When using the lexicon alone, it is more appropriate to make claims about VAD word usage rather than the VAD of the speakers. For example, {\it `the use of high-valence words in the context of the target group grew by 20\%'} rather than {\it `valence in the target group grew by 20\%'}. In certain contexts, and with additional information, the inferences from word usage can be used to make broader VAD claims. 
\item \textit{Avoid Over Claiming:} Inferences drawn from larger amounts of text are often more reliable than those drawn from small amounts of text.
 For example, {\it `the use of high-valence words grew by 20\%'} is informative when determined from hundreds, thousands, tens of thousands, or more instances. Do not draw inferences about a single sentence or utterance from the VAD associations of its constituent words.
\item \textit{Embrace Comparative Analyses:} Comparative analyses can be much more useful than stand-alone analyses. Often, VAD word counts and percentages on their own are not very useful. 
For example, {\it `the use of high-valence words grew by 20\% when compared to [data from last year, data from a different person, etc.]'} is more useful than saying {\it `on average, 5 high-valence words were used in every 100 words'}.
\end{compactenum}
\noindent We recommend careful reflection of ethical considerations relevant for the specific context of deployment when using the VAD lexicon.

% \section*{Acknowledgments}

% Many thanks to Tara Small for  helpful discussions.

%\bibliography{acl2017}
% \section{Bibliographical References}
\bibliography{maxdiff}
\bibliographystyle{acl_natbib}

\appendix

\section{APPENDIX}
\label{appendix-a}

\subsection{AMT Questionnaires for Valence, Arousal, and Dominance}
\begin{figure*}[t]
	     \centering
	     \includegraphics[width=\textwidth]{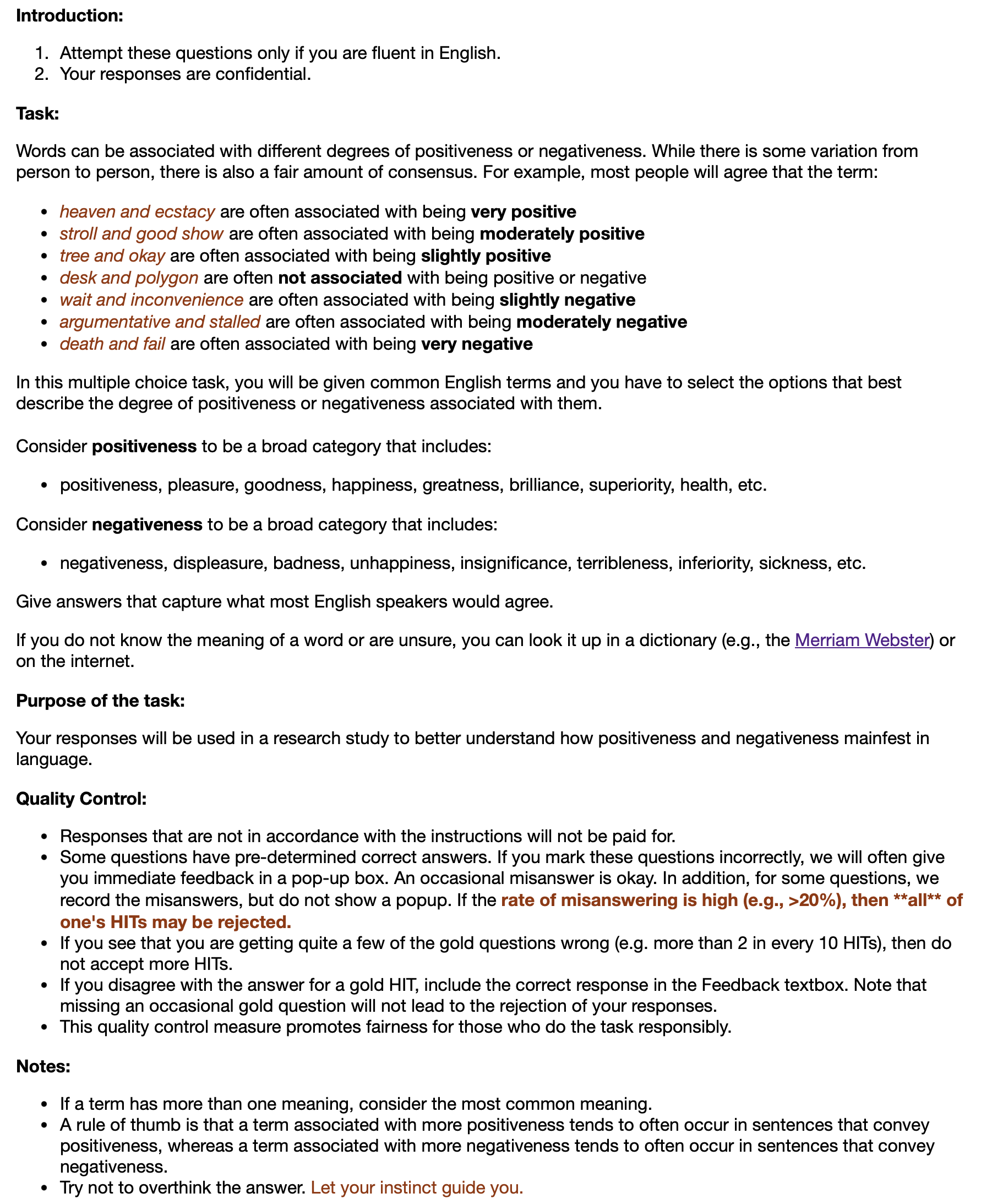}
	     \caption{Valence Questionnaire: Detailed instructions.}
	     \label{fig:val-q}
	 \end{figure*}

\begin{figure*}[t]
	     \centering
	     \includegraphics[width=\textwidth]{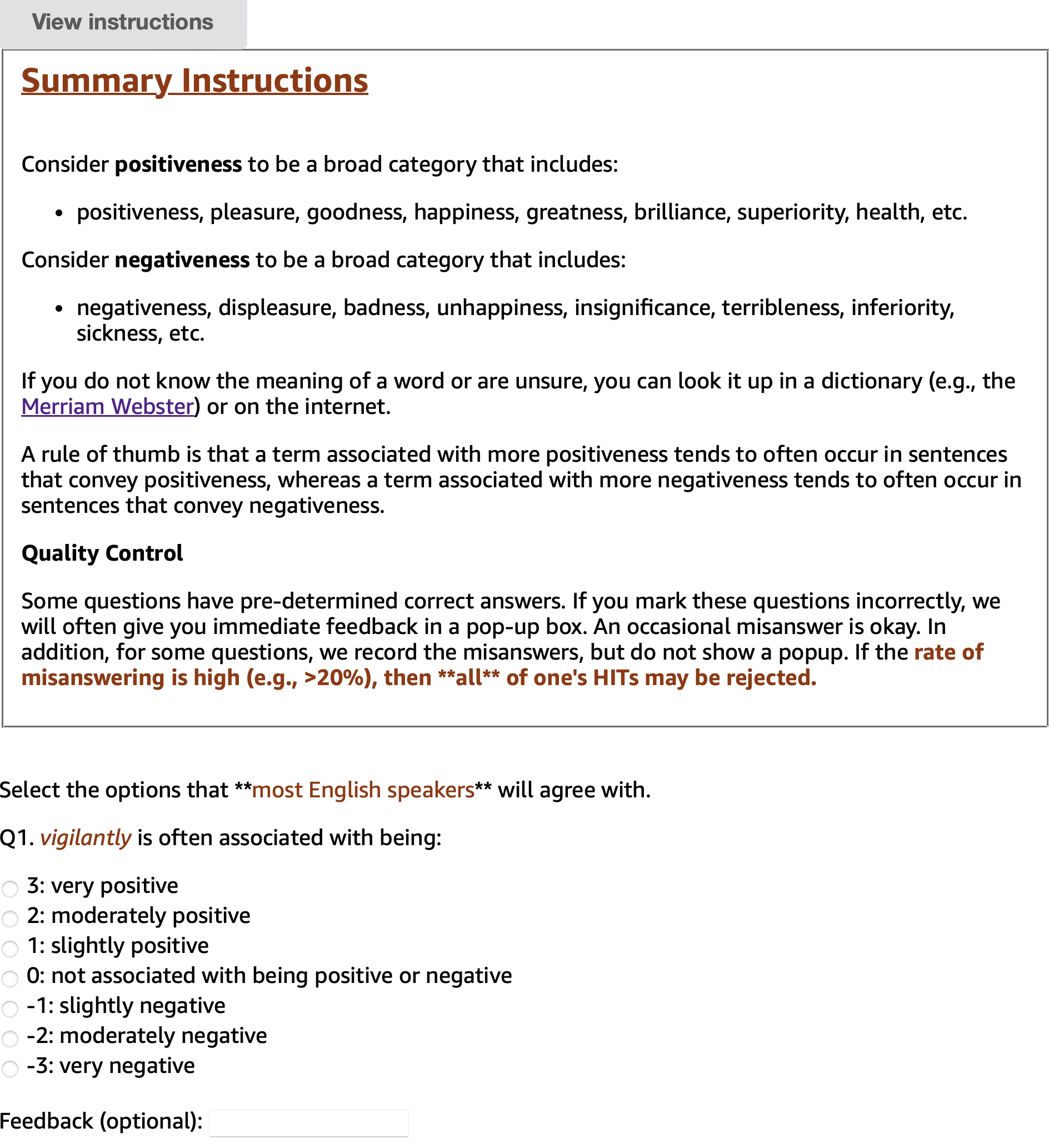}
	     \caption{Valence Questionnaire: Sample question.}
	     \label{fig:val-sumq}
	 \end{figure*}

\begin{figure*}[t]
	     \centering
	     \includegraphics[width=0.65\textwidth]{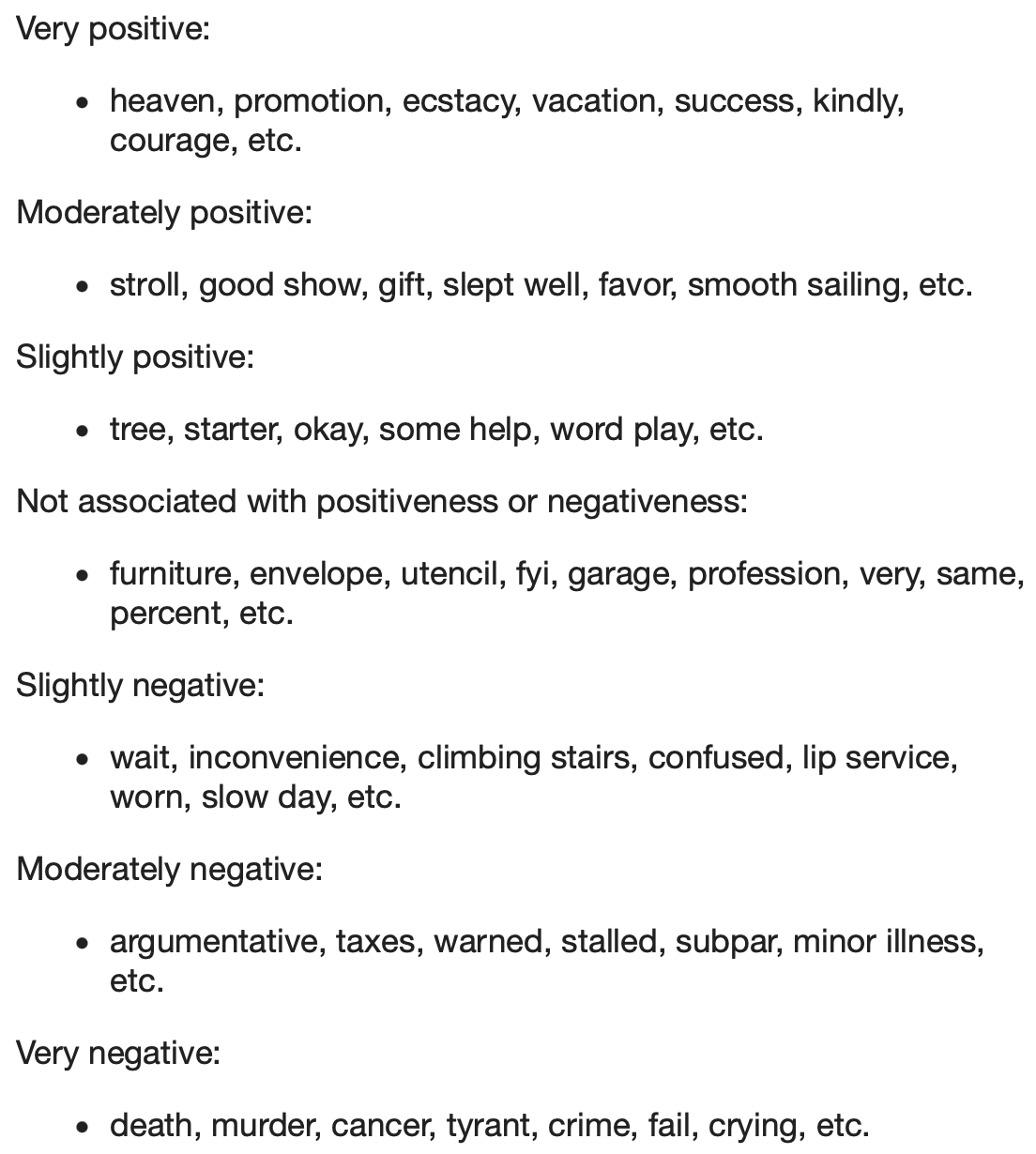}
	     \caption{Valence Questionnaire: Examples.}
	     \label{fig:val-ex}
	 \end{figure*}

\begin{figure*}[t]
	     \centering
	     \includegraphics[width=\textwidth]{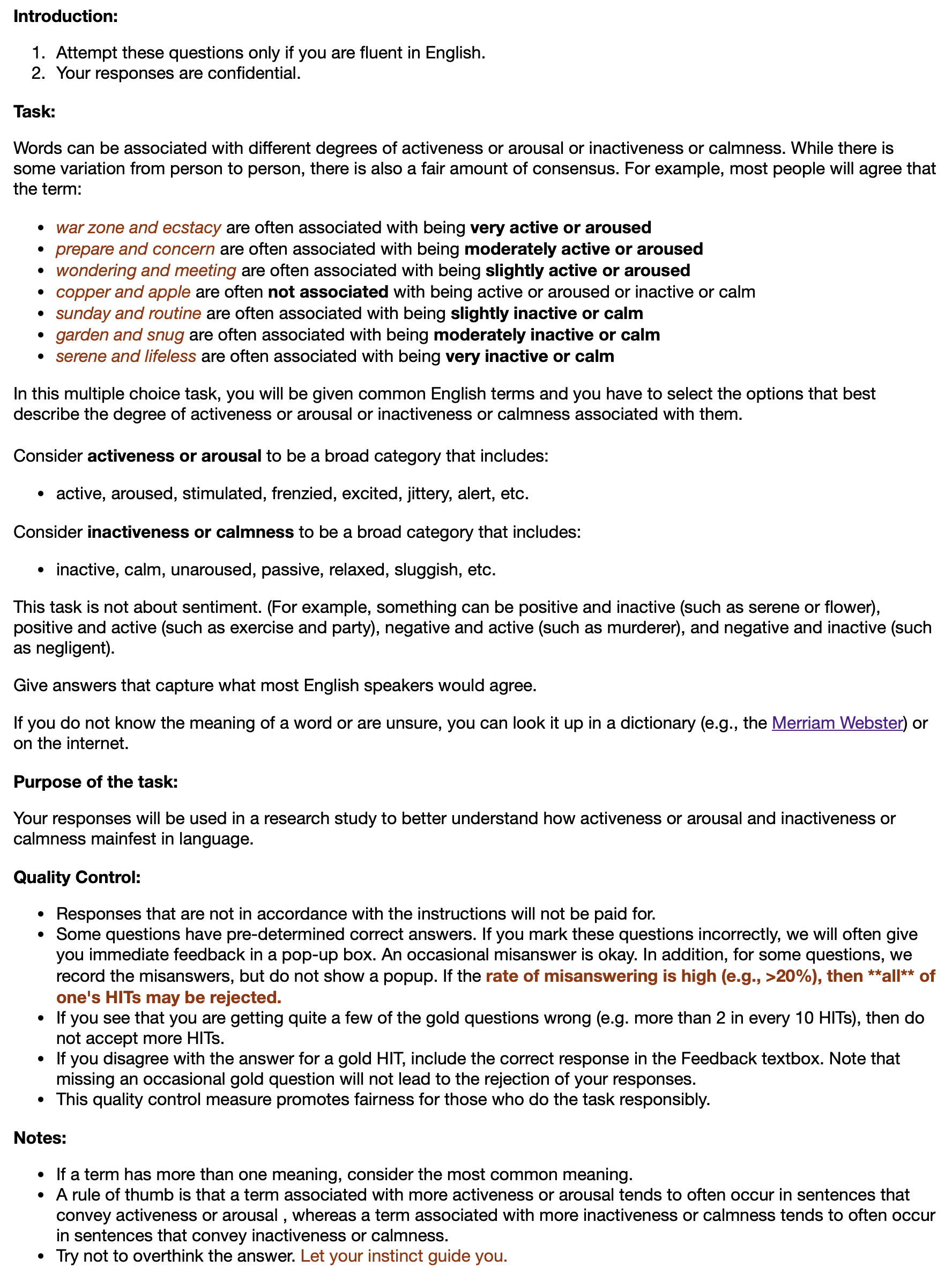}
	     \caption{Arousal Questionnaire: Detailed instructions.}
	     \label{fig:aro-q}
	 \end{figure*}

\begin{figure*}[t]
	     \centering
	     \includegraphics[width=\textwidth]{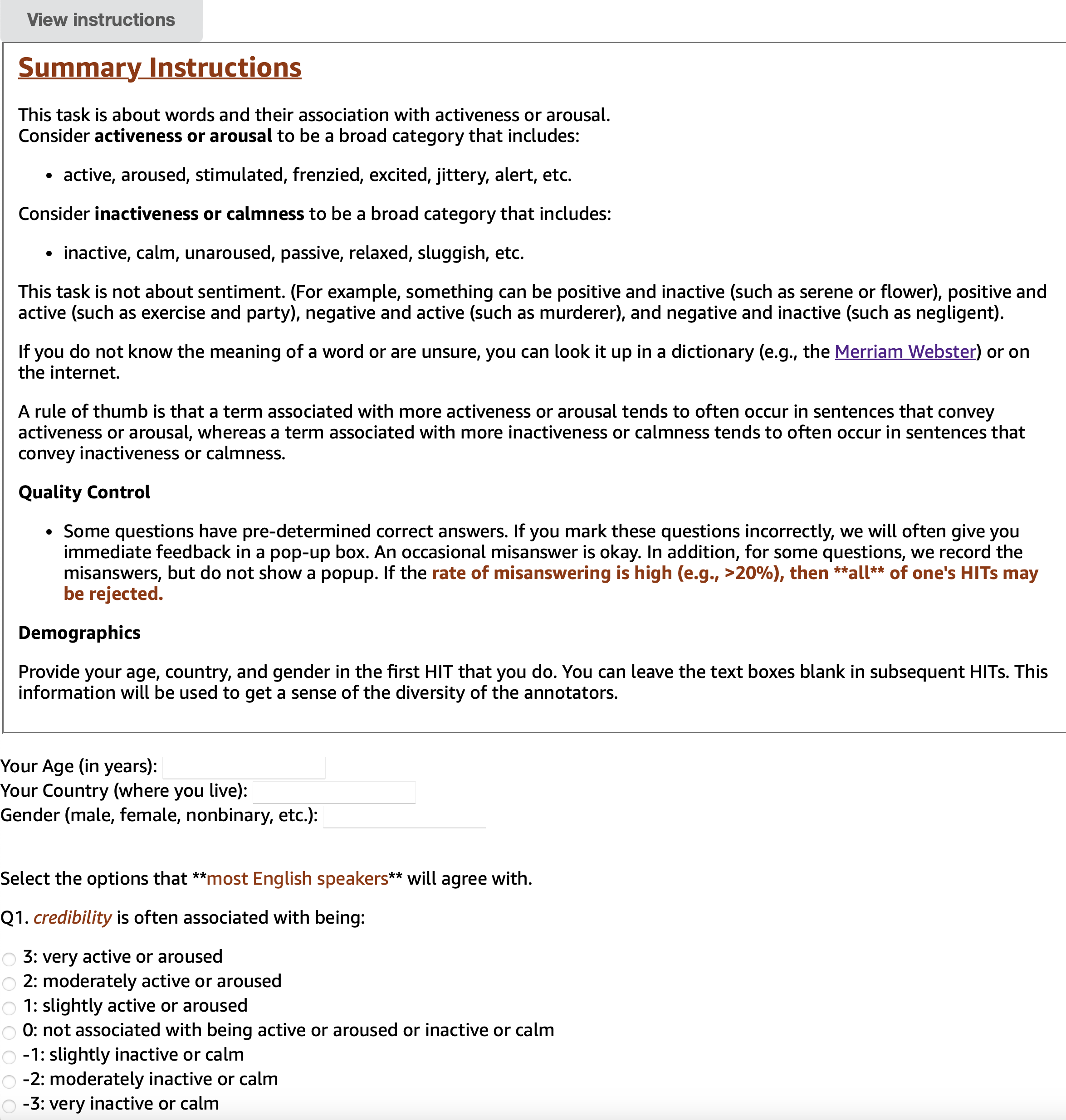}
	     \caption{Arousal Questionnaire: Sample question.}
	     \label{fig:aro-sumq}
	 \end{figure*}

\begin{figure*}[t]
	     \centering
	     \includegraphics[width=0.75\textwidth]{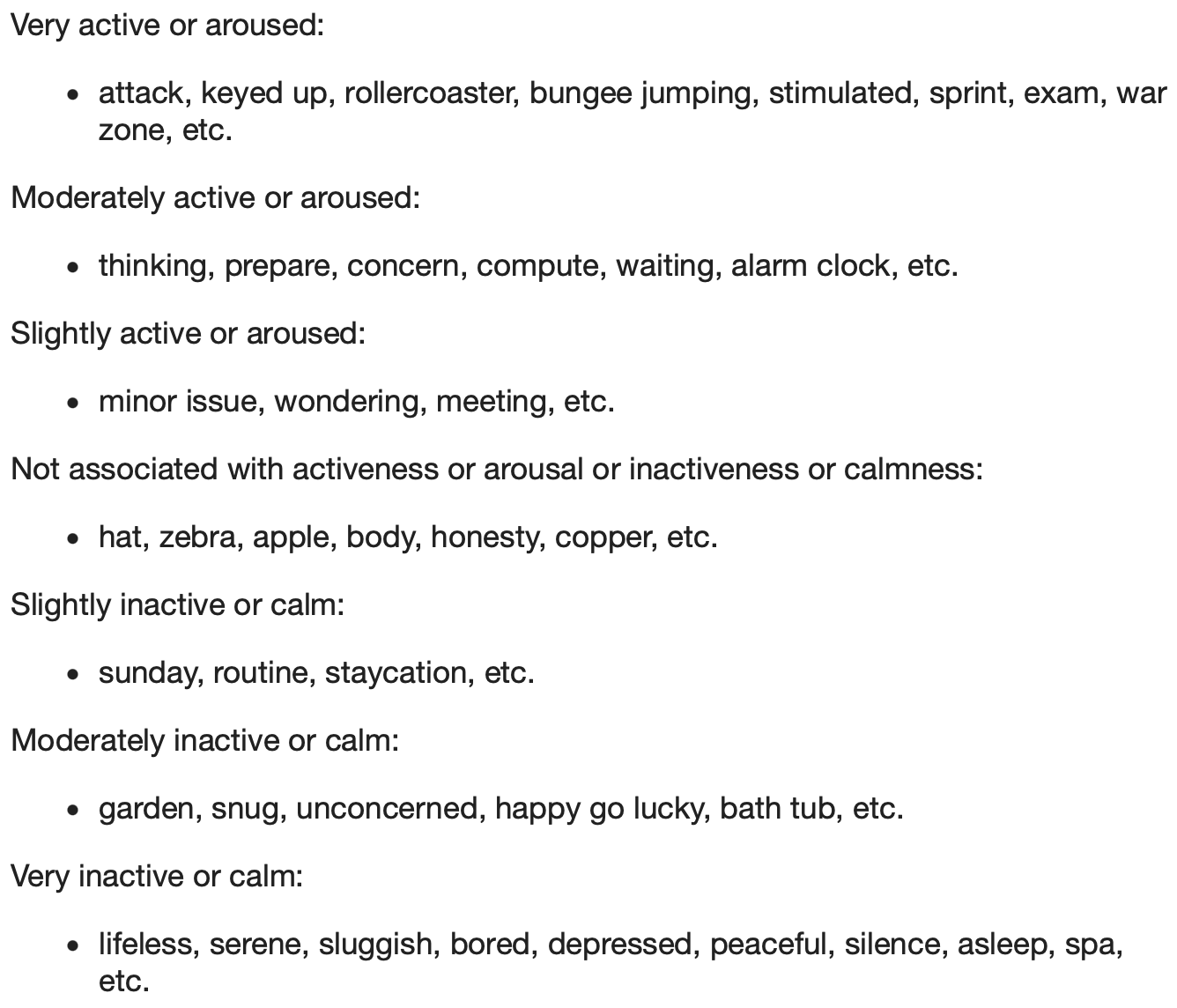}
	     \caption{Arousal Questionnaire: Examples.}
	     \label{fig:aro-ex}
	 \end{figure*}

\begin{figure*}[t]
	     \centering
	     \includegraphics[width=\textwidth]{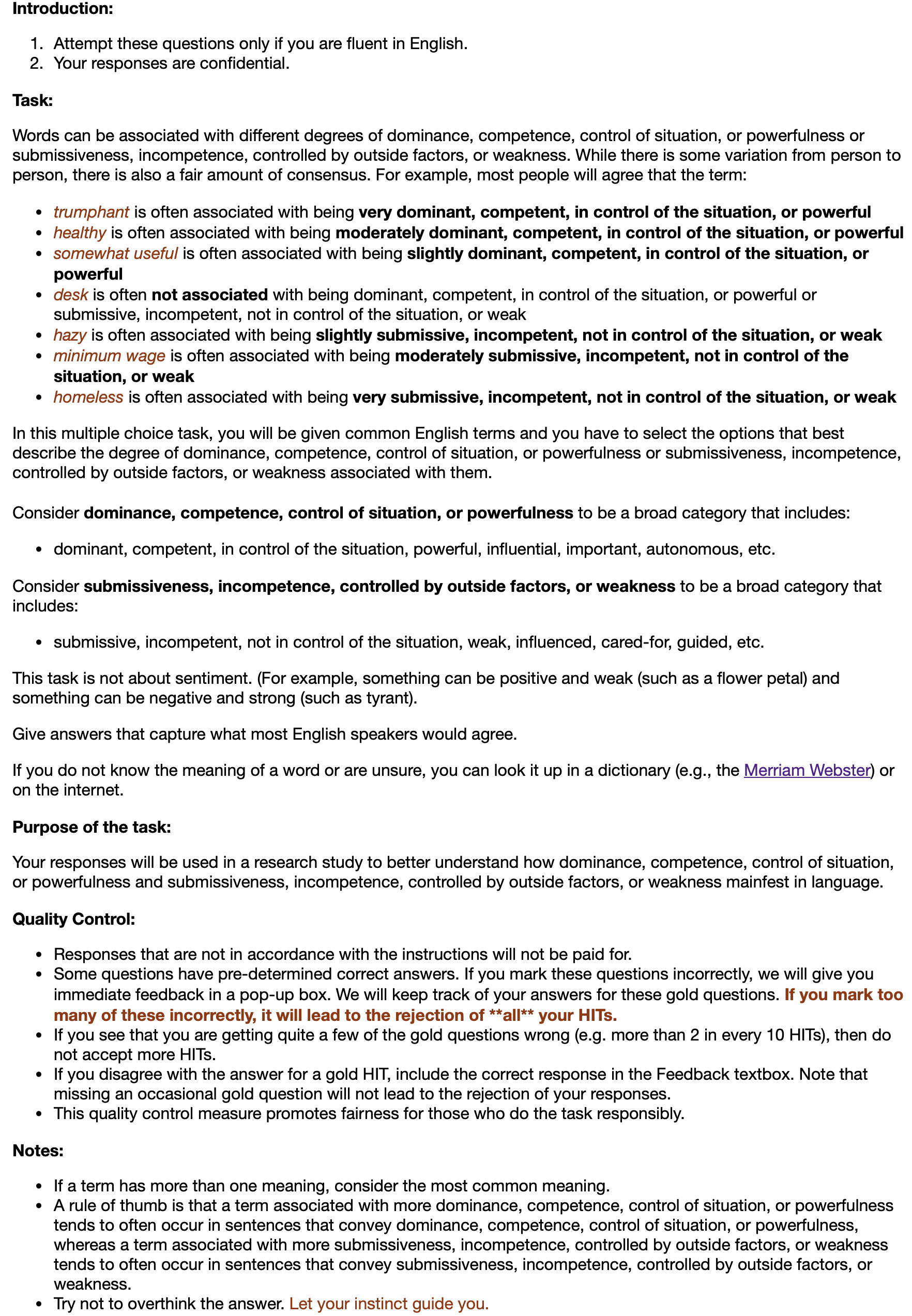}
	     \caption{Dominance Questionnaire: Detailed instructions.}
	     \label{fig:dom-q}
	 \end{figure*}

\begin{figure*}[t]
	     \centering
	     \includegraphics[width=\textwidth]{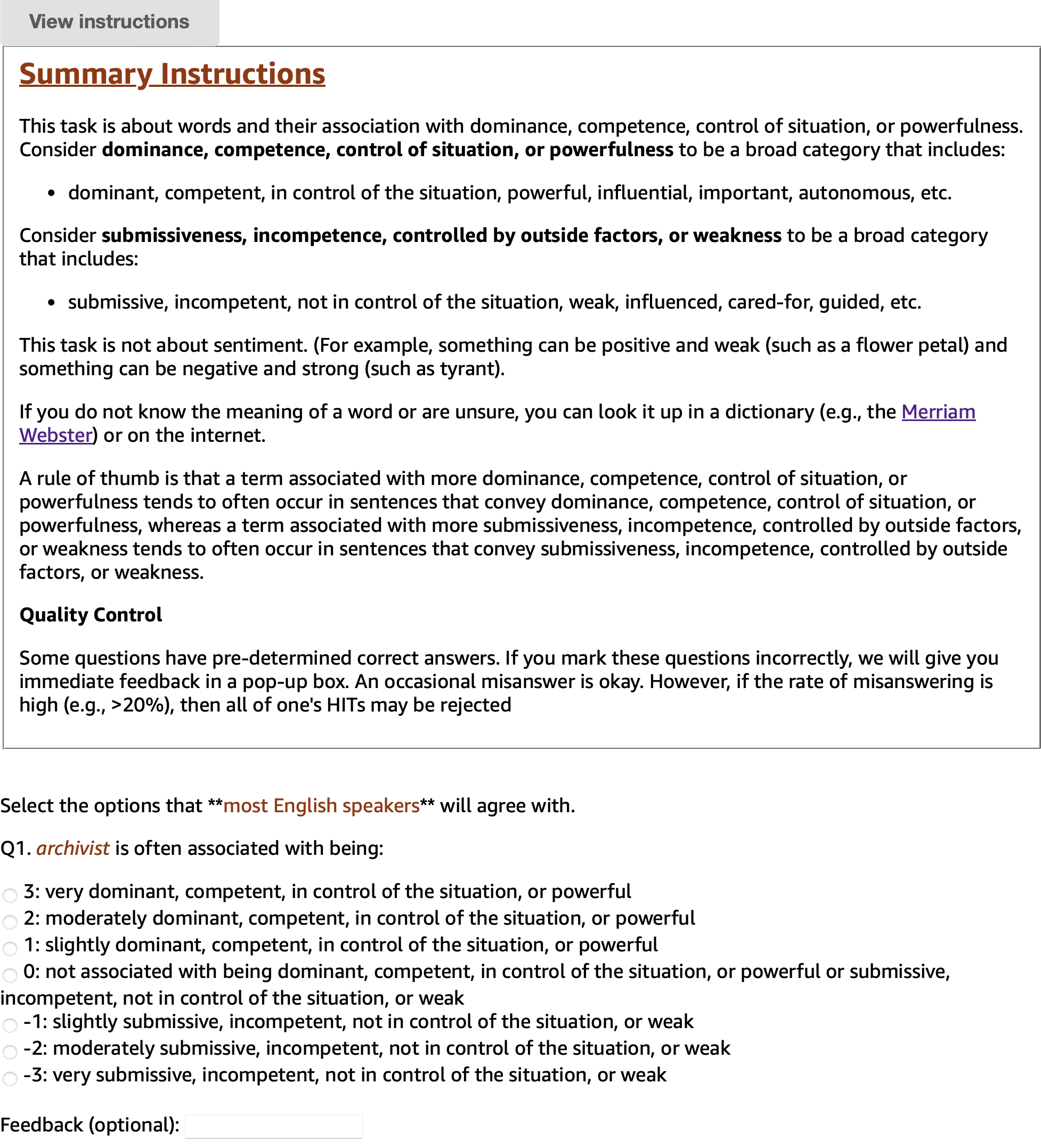}
	     \caption{Dominance Questionnaire: Sample question.}
	     \label{fig:dom-sumq}
	 \end{figure*}

\begin{figure*}[t]
	     \centering
	     \includegraphics[width=0.75\textwidth]{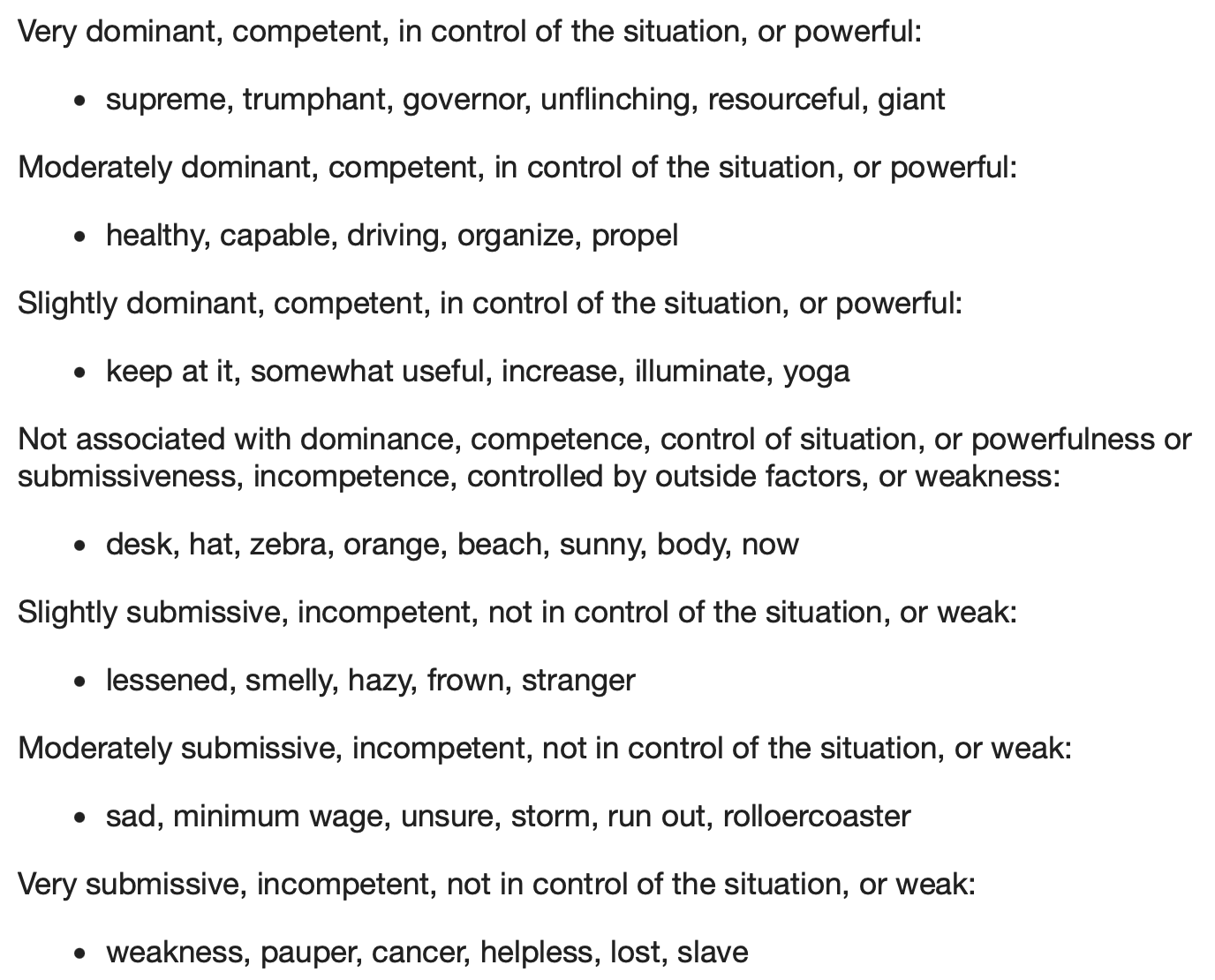}
	     \caption{Dominance Questionnaire: Examples.}
	     \label{fig:dom-ex}
	 \end{figure*}

Screenshots of the detailed instructions, sample instance (question), and examples presented to the annotators are shown in Figures \ref{fig:val-q} through \ref{fig:dom-ex}. Participants were informed that they may work on as many instances as they wish. 

% ----------------

% ----------------

\end{document}